\documentclass[manuscript,screen]{acmart}

\usepackage{tikz,pgfplots,pgfplotstable}
\usepackage{float}
\usepackage{caption} 
\usepackage{subcaption}
\usepackage{graphicx}
\usepackage{multicol}

\pgfplotsset{compat=default}
\usetikzlibrary{pgfplots.groupplots}

\pgfplotsset{compat=1.11,
    /pgfplots/ybar legend/.style={
    /pgfplots/legend image code/.code={%
       \draw[##1,/tikz/.cd,yshift=-0.25em]
        (0cm,0cm) rectangle (3pt,0.8em);},
   },
}

\pgfkeys{/pgf/number format/.cd,
    fixed,
    set thousands separator={}}
    
\AtBeginDocument{%
  \providecommand\BibTeX{{%
    \normalfont B\kern-0.5em{\scshape i\kern-0.25em b}\kern-0.8em\TeX}}}

\usetikzlibrary{datavisualization.formats.functions}

\setcopyright{acmcopyright}
\copyrightyear{2023}
\acmYear{2023}
\acmDOI{}

\acmConference[EAAMO '23]{ACM Conference on Equity and Access in Algorithms, Mechanisms and Optimization}{October -- 2023}{Boston, US}
\acmPrice{}
\acmISBN{}

\begin{document}

\title{Towards a Better Understanding of the Computer Vision Research Community in Africa}

\author{Abdul-Hakeem Omotayo}
\authornote{Equal contribution - we chose to write the country of nationality (.) in addition to the country of the affiliated institution if different, as we are proud of both our current institutions and our African background. Correpsonding e-mail: roya.cv4africa@gmail.com}
\email{ormorteey@gmail.com}
\affiliation{%
  \institution{University of California, Davis}
  \country{US, (Nigeria)}
}

\author{Mai Gamal}
\authornotemark[1]
\email{mai.tharwat@guc.edu.eg}
\affiliation{%
  \institution{German University in Cairo}
  \country{Egypt}
}

\author{Eman Ehab}
\authornotemark[1]
\email{e.ehab@nu.edu.eg}
\affiliation{%
  \institution{Nile University}
  \country{Egypt}
}

\author{Gbetondji Dovonon}
\authornotemark[1]
\email{gbetondji.dovonon.22@ucl.ac.uk}
\affiliation{%
  \institution{University College London}
  \country{UK, (Benin)}
}

\author{Zainab Akinjobi}
\authornotemark[1]
\email{akinzayn@gmail.com}
\affiliation{%
  \institution{New Mexico's State University}
  \country{US, (Nigeria)}
}

\author{Ismaila Lukman}
\authornotemark[1]
\email{ismailukman@gmail.com}
\affiliation{%
  \institution{University of Angers}
  \country{France, (Nigeria)}
}

\author{Houcemeddine Turki}
\authornotemark[1]
\email{turkiabdelwaheb@hotmail.fr}
\affiliation{%
  \institution{University of Sfax}
  \country{Tunisia}
}

\author{Mahmod Abdien}
\authornotemark[1]
\email{21mmah@queensu.ca}
\affiliation{%
  \institution{Queen's University}
  \country{Canada, (Egypt)}
}

\author{Idriss Tondji}
\authornotemark[1]
\email{itondji@aimsammi.org}
\affiliation{%
  \institution{African Master's in Machine Intelligence (AMMI-AIMS)}
  \country{Senegal, (Cameroon)}
}

\author{Abigail Oppong}
\authornotemark[1]
\email{abigoppong@gmail.com}
\affiliation{%
  \institution{Ashesi University}
  \country{Ghana}
}

\author{Yvan Pimi}
\authornotemark[1]
\email{ypimi@aimsammi.org}
\affiliation{%
  \institution{African Master's in Machine Intelligence (AMMI-AIMS)}
  \country{Senegal, (Cameroon)}
}

\author{Karim Gamal}
\authornotemark[1]
\email{21kgmm@queensu.ca}
\affiliation{%
  \institution{Queen's University}
  \country{Canada, (Egypt)}
}

\author{Ro'ya}
\email{roya.cv4africa@gmail.com}
\affiliation{%
  \institution{Computer Vision for Africa Grassroots}
  \country{Africa}
  }

\author{Mennatullah Siam}
\email{mennatullah.siam@ontariotechu.ca}
\affiliation{%
  \institution{Ontario Tech University}
  \country{Canada, (Egypt)}
}

\renewcommand{\shortauthors}{Ro'ya, et al.}

\begin{abstract}
 Computer vision is a broad field of study that encompasses different tasks (e.g., object detection, semantic segmentation, 3D reconstruction). Although computer vision is relevant to the African communities in various applications, yet computer vision research is under-explored in the continent and constructs only 0.06\% of top-tier publications in the last ten years. In this paper, our goal is to have a better understanding of the computer vision research conducted in Africa and provide pointers on whether there is equity in research or not. We do this through an empirical analysis of the African computer vision publications that are Scopus indexed, where we collect around 63,000 publications over the period 2012-2022. We first study the opportunities available for African institutions to publish in top-tier computer vision venues. We show that African publishing trends in top-tier venues over the years do not exhibit consistent growth, unlike other continents such as North America or Asia. Moreover, we study all computer vision publications beyond top-tier venues in different African regions to find that mainly Northern and Southern Africa are publishing in computer vision with 68.5\% and 15.9\% of publications, resp. Nonetheless, we highlight that both Eastern and Western Africa are exhibiting a promising increase with the last two years closing the gap with Southern Africa. Additionally, we study the collaboration patterns in these publications to find that most of these exhibit international collaborations rather than African ones. We also show that most of these publications include an African author that is a key contributor as the first or last author. Finally, we present the most recurring keywords in computer vision publications per African region. In summary, our analysis reveals that African researchers are key contributors to African research, yet there exists multiple barriers to publish in top-tier venues and establish African collaborations. Additionally, the question on whether there is an alignment between the current computer vision topics published in Africa and the most urgent needs in African communities remains unanswered. In this work we took the first step of documenting per-region published topics and we leave it for future work to investigate the latter question. This work is part of a community based effort that is focused on improving the computer vision research in Africa, where we question whether researchers across the different regions have access to equal opportunities to lead their research or not.
\end{abstract}





\received[accepted]{28 August 2023}

\maketitle

\section{Introduction}

Computer vision is a sub-field of artificial intelligence that enables computers to interpret and derive information from the visual world represented in images or videos with different sub-tasks (e.g., object recognition, reconstruction, and re-organization). It is relevant in various applications especially in Africa such as remote sensing~\cite{sefala2021constructing,sirko2021continental}, medical image processing~\cite{manescu2020weakly,nakasi2020web,roshanitabrizi2022ensembled}, robotics (e.g, understanding ego-centric videos~\cite{grauman2022ego4d}) and AI fairness~\cite{kinyanjui2020fairness,buolamwini2018gender}. Yet, we show in this work that most of the computer vision research in top-tier venues is happening outside Africa, even when Africans are the lead researchers or contributing to it. Previous studies have discussed and documented biases in some of the existing artificial intelligence methods~\cite{mehrabi2021survey,obermeyer2019dissecting,sweeney2013discrimination}, especially in computer vision~\cite{buolamwini2018gender,kinyanjui2020fairness}. Recently, this is even more relevant with the emergence of foundation models~\cite{bommasani2021opportunities} that are trained on broad data and adapted to a multitude of tasks, but are encoding and amplifying biases~\cite{bender2021dangers}.  However, these biases stem originally from the fact that historically marginalized populations are not sufficiently consulted or put to be part of the design of these methods, training datasets, or design of the computer vision systems~\cite{mohamed2020decolonial}. Recently, more works have discussed a decolonial approach to artificial intelligence that empowers these historically marginalized populations~\cite{lewis2020indigenous,mhlambi2020rationality,mohamed2020decolonial,bondi2021envisioning}.

A discussion on the different participatory grassroots efforts that empower historically marginalized communities has been previously documented~\cite{fourie2023a}. Communities such as Black in AI (BAI), Data Science Nigeria (DSN), Deep Learning Indaba (DLI), Masakhane, and Sisonke Biotik all have contributed to empowering African researchers. These communities and their events have had positive and noticeable impact on the machine learning community in Africa. However, there is minimal focus on the computer vision field and most of the events of the aforementioned communities happen at machine learning conferences such as the conference on \emph{Neural Information Processing Systems (NeurIPS)}. Looking at the top ten most cited researchers in computer vision, none of them is from an African background to the best of our knowledge~\cite{top50}. Moreover, Looking at widely acknowledged sources discussing the computer vision history~\cite{szeliski2022computer}, it mostly focuses on the recent trends in deep learning, vision and language models. It misses discussions on fairness in machine learning and computer vision, except for a single citation discussing the emotional impact of working on computer vision research~\cite{su2021affective}. Although, it explicitly states that, it is the authors' own views, yet it mostly reflects developed countries' view of what is considered impactful research. From the perspective of historically marginalized communities, AI fairness has a huge impact on their lives equal or even more than what vision-language models can offer them. Hence, why it is important to empower the computer vision research ecosystem in these communities so they can lead their own research to bring new perspectives that are generally undermined in the field. Mohamed \textit{et. al.}~\cite{mohamed2020decolonial} discussed algorithmic dispossession in which marginalized communities are designed for instead of participating in the design, and proposed three steps toward dismantling it. These include the reciprocal engagements between the affected communities and researchers and active engagement with communities and grassroots organizations. This work is part of a grassroots computer vision Ro'ya\footnote{\url{https://ro-ya-cv4africa.github.io/homepage/}} that aims at bridging the current gap in African grassroots.

In this paper, we propose to do a bibliometric study of African computer vision Scopus indexed publications, followed by a summary of insights and discussion of the current state in African computer vision research. We do this towards a better understanding of the research landscape in Africa in terms of strengths, weaknesses, and barriers facing the researchers inspired by previous work~\cite{turki2023machine}. We aim to document the discrepancies between African computer vision publications in top-tier venues against other venues. Although, some might argue that on the road to decolonizing computer vision, we should not be tied to top-tier venues to assess success. However, it is important to keep in mind that the merits of publishing in top-tier venues such as, better scholarships, fellowships, grants, and better visibility and impact for the work, necessitates such study. This is in part towards ensuring equal opportunity and access to conduct such research within African institutions. Moreover, in this study we aim to understand the geo-temporal trends of publishing in different African regions, the publishing patterns and collaborations with non-African institutions, and whether African authors are key contributors to these publications. Our study also includes an analysis of the keywords used to better understand what is currently happening in computer vision research in Africa. Finally, we go beyond reporting statistics only and convey the community members' experience, to shed light on some of the barriers we face. Our contributions are as follows:
\begin{itemize}
\item We present the first bibliometric study on African computer vision research with the goal of a better understanding of our research landscape and to document inequity in computer vision research.
\item We document discrepancies between publishing in top-tier and other computer vision venues and demonstrate a novel way to use the affiliation history towards this end. The affiliation history of the researchers is used as a proxy to identify some of the African researchers publishing in non-African institutions.
\item We analyze the publishing temporal trends per African region, the collaboration patterns and the most frequent keywords/topics distribution per region, followed by a discussion on the equity in computer vision research.
\end{itemize}

\section{Related Works}

\textbf{Bibliometric studies:} In the related scientometric and bibliometric studies there exists multiple works that have studied scientific publications generally from Africa~\cite{pouris2014research,sooryamoorthy2021science} or with respect to a specific topic~\cite{tlili2022we,guleid2021bibliometric,musa2022bibliometric} such as health sciences~\cite{musa2022bibliometric} or COVID~\cite{guleid2021bibliometric}. One of the earliest studies~\cite{pouris2014research} showed that African countries mostly focus on international collaborations rather than collaborating within the continent. They highlight that these are mainly driven by the availability of resources and interests outside Africa. We think that encouraging these international collaborations is still an important component in improving African research. However, it is more important to empower the African research ecosystem to have some form of independence and increase the collaborations within the continent, while being connected to international collaborators. This is especially evident in computer vision, which is largely driven nowadays by large-scale datasets and models that are trained on expensive compute. Thus, we encourage international collaborations without undermining an independent African research that focuses on our own problems and needs. 

A closely related work~\cite{turki2023machine} studied African publications in machine learning for health. Their main findings included that Northern African countries had the most substantial contributions with respect to other African regions. However, this trend reduced over the years with more contributions emerging from sub-Saharan Africa. It also confirmed the correlation between international funding and collaborations on increasing the contributions from Africa. Inspired by this past work, we perform a bibliographic analysis of the African contributions, but we rather focus on the computer vision field which is more diverse with various applications from medical image processing to remote sensing. It is worth noting that previous bibliometric studies lacked the focus on the computer vision field, which we argue is urgent nowadays to Africa. In the computer vision field, some bibliometric studies focused on certain topics like convolutional neural networks~\cite{chen2020bibliometric} or specific applications~\cite{iqbal2023last}. Yet, these do not focus on the African context which is our key question towards a decolonial computer vision approach that we encourage within our participatory framework.




\textbf{Grass-roots participatory framework:} Recent work~\cite{bondi2021envisioning} has discussed a critique of the definition of AI for social good and how to evaluate the project's goodness. They propose a PACT framework; a Participatory approach to enable capabilities in communities; to achieve this goal. In this framework, they provide a list of guiding questions that can help researchers assess the goodness in the project within a participatory framework. This relates to calls for a decolonial AI that was recently spread~\cite{birhane2020towards,kalluri2021don,lewis2020indigenous,mohamed2020decolonial,whittaker2019disability}, where one of these calls iterated on the importance of participatory and community based efforts~\cite{mohamed2020decolonial}. Towards a participatory grassroots framework~\cite{fourie2023a} a discussion comparing top-down \textit{vs.} bottom-up approaches with the description of the types of grassroots communities that emerged recently was presented. These include: (i) affinity based organizations such as Women in Machine Learning and Black in AI, (ii) topic based communities such as Masakhane, and Sisonke Biotik, and (iii) event based ones such as Deep Learning Indaba. The framework they proposed is focused on African grassroots, where they discuss the common values and participation roles within such communities. One specific community, Sisonke Biotik, is focusing on machine learning for health. Their initial project was a bibliometric study of African research in machine learning for health~\cite{turki2023machine}. Inspired by such efforts, we aim to perform a similar study focused on the computer vision field. 

However, unlike previous works, we discuss publications in top-tier venues and African contributions there to document quantitatively the equity in research and assess Africans' access to opportunities in the field. Top tier publications could be an entry point for a lot of opportunities in terms of scholarships, research grants, and collaborations. It can impact the kind of datasets and compute that African researchers can have access to. Moreover, unlike previous works, we do not limit our study to quantitative analysis only. Although unusual in bibliometric studies, we reflect on some of the community members' stories in the discussion as part of our commitment to \textit{``community first''}. We plan to expand this later to a quantitative survey among African computer vision researchers.

\section{Methods}
\label{sec:method}

\begin{figure*}[t]
    \centering
       \includegraphics[width=\textwidth]{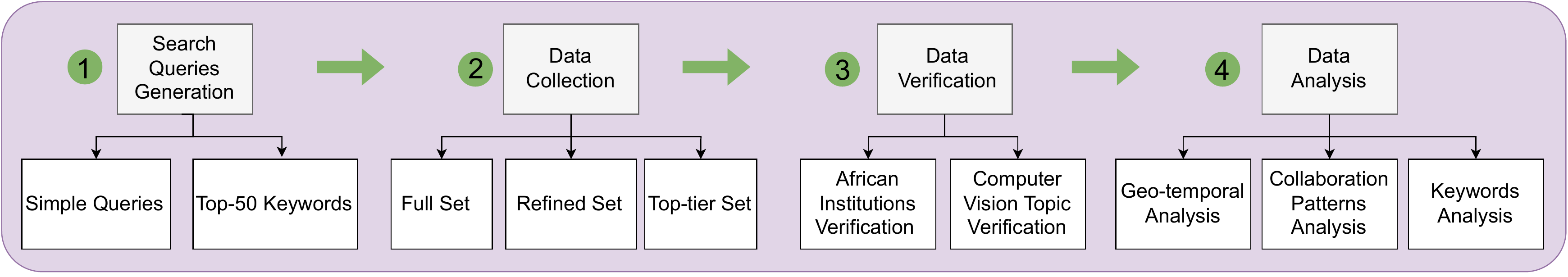}
    \caption{Our proposed pipeline for data collection, verification and analysis of the African Scopus indexed computer vision publications. The search query generation uses simple queries to retrieve all computer vision publications (i.e., \emph{full} set) or generates queries based on the Top-50 keywords in computer vision as a publications sample (i.e., \emph{refined} set). This is followed by data collection of the \emph{full}, \emph{refined} and \emph{top-tier} publications sets and a verification phase on the \emph{refined} and \emph{top-tier} sets. Finally, we present three different types of analysis on the these data sources.}
    \label{fig:methodfig}
\end{figure*}

In this section, we detail our method for gathering the necessary data for our bibliometric study with focus on Scopus indexed publications. Figure~\ref{fig:methodfig} describes four main stages: (i) automatic search query generation, (ii) data collection, (iii) data verification and (iv) data analysis. The computer vision field is quite broad and multidisciplinary, it overlaps with machine learning for health but also has a broader set of applications beyond this. We aim for collecting three types of data. The first type is African publications that are Scopus indexed with any relevance to computer vision in general, we refer to this set as the \emph{full} data. The second type is a reduced set of African publications that focuses on the top 50 keywords used in the computer vision field, we refer to this set as the \emph{refined} data. Finally, we collect the publications in top tier venues in the computer vision field as the \emph{top-tier} set. In the following, we will describe the details of the aforementioned four stages and the collection of these three sets of data. 

\subsection{Query generation and data collection}
In the \emph{full} publications set, where we aim to collect all African publications relevant to the computer vision field, we restrict our search query to African countries and the search keyword to \emph{(``image'' OR ``computer vision'')}. Publications with at least one author from an African institution are the ones considered. We also restrict the time interval from 2012 to 2022 with the start of the deep learning era where convolutional neural networks won the 2012 ImageNet~\cite{deng2009imagenet} challenge~\cite{krizhevsky2017imagenet}. The \emph{full} publications set is approximately 63,000 publications. Due to the large-scale nature of this set, we do not verify it but we still use it to provide insights on the general trends in African computer vision.

As for the \emph{refined} publications set, we only focus on the top 50 keywords used in computer vision publications to reduce our data and allow for a consecutive verification phase. This set includes approximately 18,000 publications where the search query was restricted to African countries similar to the \emph{full} set. The search keyword used is \emph{(``image'' OR ``computer vision'') AND (KEYWORD)}, where \emph{KEYWORD} was replaced with noun phrases from the top 50 computer vision keywords that include \emph{``deep learning'', ``object detection'', ``image segmentation'', ``robotics''} as examples. The top 50 keywords were retrieved using an off-the-shelf tool as described later. This refined and reduced publications set allowed for a verification phase in order to remove false positives. False positives stem from being wrongly assigned as relevant to computer vision or having at least one African author. 

Finally, for the \emph{top-tier} publications set we focus on top-tier conferences and journals that are well-acknowledged in the computer vision field without any geographical restrictions. We use the CORE\footnote{\url{http://portal.core.edu.au/conf-ranks/}} system to identify rank $\text{A}^*$ computer vision venues and few of the machine learning ones that include computer vision publications. The top-tier conferences we use include: \emph{CVPR, ICCV, ECCV, ICML, ICLR, NeurIPS} and we add \emph{MICCAI} as the top-tier conference with medical image processing publications. For the top-tier journals, we focus on \emph{TPAMI, IJCV}. We restrict the time interval from 2012 to 2022. The final \emph{top-tier} set has approximately 43,000 publications. We do acknowledge that some of the top-tier publications in \emph{ICML, ICLR, NeurIPS} can be on general machine learning but we see it still as a good statistic on understanding how African institutions contribute to these venues. Moreover, we provide a separate analysis on the most frequent and established computer vision venue (i.e., \emph{CVPR}) to avoid the aforementioned issue. We also perform a verification on this set, to remove false positives in terms of publication types such as ``Retracted'' or ``Review''.

Throughout the data collection stage for the \emph{refined} and \emph{top-tier} sets we use Scopus APIs\footnote{\url{https://pybliometrics.readthedocs.io/en/stable}}. We mildly use SciVal\footnote{\url{https://scival.com/home}} in the selection of the top-50 keywords and in the \emph{full} set collection, as it has constrained usage. For example, it has limited control on the time-interval and the retrieved meta-data such as the authors' countries and affiliation history.

\subsection{Data verification and analysis}
\label{sec:method_details}
Previous bibliometric studies rarely choose to perform a verification phase and rather analyze directly the collected publications set from different web sources. A verification phase was conducted in some bibliometric studies, but outside the African context, and they were restricted to small publication sets up to 3,000 publications~\cite{young2018characterizing,jun2022most}. To the best of our knowledge, we are the first to design a verification phase within the African bibliometric studies literature with the scale of 18,000 publications. Our goal is to reduce sources of errors to have a better understanding of African computer vision research. It is even more important when working on such a diverse topic as computer vision. The verification phase includes a combination of automatic and manual verification that we will describe in detail. Since the \emph{full} set is around 63,000 entries we find it difficult to verify and we rather focus on the \emph{refined} and \emph{top-tier} sets. 

The reason we chose to perform an initial verification phase as we found three sources of errors in the \emph{refined} and \emph{top-tier} sets. The sources of errors include: (i) African authorship errors, (ii) computer vision relevance issues and (iii) irrelevant publication types. We here present some examples on these sources of errors and issues. For the African authorship errors, one example is labelling Papua New Guinea as an African country conflating it with Guinea which was identified in multiple publications~\cite{Sirohi2021487,khamparia2020classification,doaemo2020exploring}. Others include typos in the country such as Swaziland instead of Switzerland that was identified in this publication~\cite{Jin20171705}. As for the computer vision relevance issues, one example we show is the use of the term ``image data'' in the abstract as an example but has no relevance to the topic being covered in the publication which operates on a simulated milling circuit dataset~\cite{McCoy2018141}. We use a broad definition of relevance to computer vision, where works that operate on image datasets to interpret these images are considered relevant to computer vision. Finally, for the publication types we filter out ``Retracted'', ``Review'', ``Erratum'', or ``Conference Review'' types, to focus on the research publications output. 

The verification phase for the computer vision relevance starts with an automatic verification by filtering out venues with less than ten publications. We put a condition that the venue name does not belong to any of the top-tier venues defined previously and does not include the keywords \emph{image} or \emph{computer vision} in the venue name. These types of venues such as \emph{Solar Energy} are less relevant to the computer vision field and we show this publication as an example~\cite{Hamed2020310}. Although this might result in false negatives where some of the relevant publications might be filtered out, we favoured a reduced false positives in the \emph{refined} set. Since the \emph{full} set covers all the data and contrasting both sets during the analysis can help us in a better understanding of the research landscape. The rest of the venues we inspect them manually to identify potential venues that could be irrelevant to the computer vision topic. For these potential venues we randomly sample publications from them to manually inspect their title and abstract to reach a conclusion on whether these publications and venues are relevant to computer vision or not. 

Additionally, we verify publications' authors and affiliations to ensure that they include at least one African institution. We start with an automatic verification on the affiliations' countries. At that stage, Papua New Guinea was identified as a false positive. This is followed by randomly sampling publications and manually inspecting the authors. 
Finally, we filter out ``Retracted'', ``Editorial'', ``Review'', ``Erratum'', ``Conference Review'' publication types to focus on research publications. The final \emph{refined} set after verification is around 12,000 publications. We retrieve the full meta data of the authors for both the \emph{refined} and \emph{top-tier} sets, and for \emph{CVPR} publications we retrieve the authors' affiliation history. To the best of our knowledge, we are the first to use the affiliation history as a proxy to retrieve African authors even if affiliated with non African institutions. Authors with at least one African institution in their history are labelled as potentially African, and manually inspected to gather 49 additional authors and their top-tier publications. 

We conduct three types of analysis on the collected datasets towards a better understanding of the African computer vision research which include: (i) A geo-temporal analysis where we take into consideration the five African regions and their publishing patterns over time. (ii) The collaboration patterns analysis where we contrast collaborations across African countries with respect to international collaborations. We also study whether African authors are key contributors as first or last authors or not. (iii) A keywords analysis where we highlight what kind of computer vision research problems are mostly tackled in the African continent per region. For the keywords analysis we use an off-the-shelf tool, SciVal\footnote{\url{http://scival.com}}, that we believe to perform standard procedure for clustering and the top three keywords for each publication is retrieved under ``Topic Name''.

\subsection{Ethical considerations}
We report here some of the ethical considerations for the dataset that we gather that can guide our future work. Since we only focus on Scopus indexed publications, there exists a bias to the English language. It is worth noting that the French language is widely used in multiple African countries, but unfortunately our dataset does not include these publications nor the ones in African languages. We also declare that our proxy to retrieve African authors working with non African institutions has some pitfalls. The affiliation history that is dated back to African institutions will only occur if these authors were able to publish back in their countries. Thus, most of the retrieved African authors using this method belong to Northern and Southern regions, which are the highest publishing regions. Finally, we document that our research team includes researchers from Nigeria, Cameroon, Benin, Egypt, Tunisia, and Ghana spanning three African regions aligning with our goals to improve equity in research within Africa. 


\section{Results and discussion}
In this section we report the results of the refined set manual validation, then we describe the results of our three analysis types: (i) a geo-temporal analysis where we study different African regions' publishing patterns over the last ten years, (ii) a collaboration patterns analysis, and (iii) a keywords analysis. Throughout all the analysis we follow~\cite{african_regions} in defining the five African regions. We choose to conduct a fine-grained study per region instead of comparing Northern \textit{vs.} sub-Saharan Africa, in order to assess the equity in computer vision research within the continent itself.

\subsection{Manual validation results}

We start with verifying the results  of our \emph{refined} set, in terms of the topic, that it belongs to computer vision, and the authors, that they include at least one African author. We randomly select 5\% of the refined set of publications, that were manually verified by the team. Each paper was assigned two annotators. For the topic verification, we manually inspected the abstract, as for the authors we ensured at least one African institution existed in the affiliations. Our results for the topic verification show 91.1\% accuracy with only 8.9\% of the publications wrongly assigned as computer vision. In the authors verification, we report 98.4\% accuracy, where it demonstrates that most of our errors were from irrelevant topics rather than the authors. Moreover, we report high inter-annotator agreement of 93.9\% and 96.7\% in both the topic and the authors verification, resp.

\subsection{Geo-temporal analysis}
\label{sec:geotemporal_analysis}

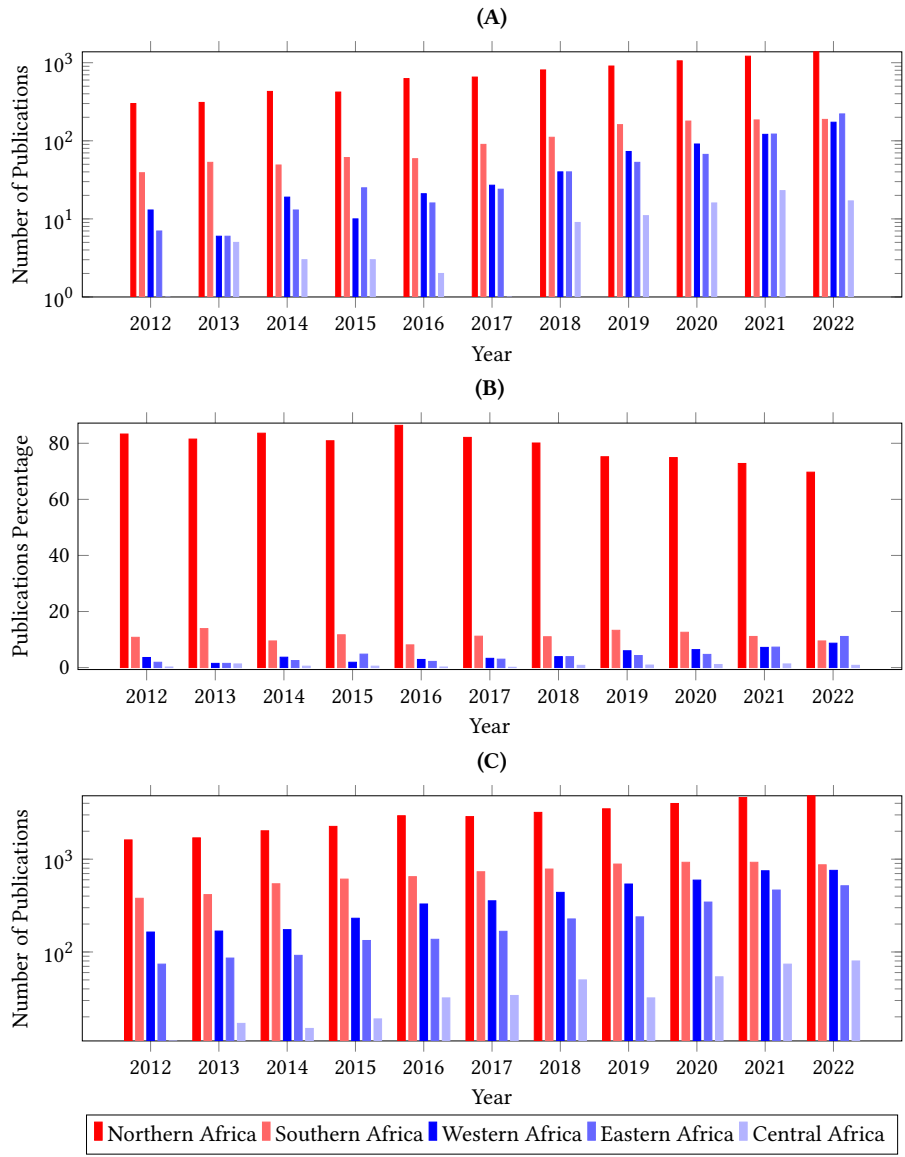
\begin{figure}
\centering
\resizebox{0.8\textwidth}{!}{
\begin{tikzpicture}
\begin{axis} [
    width=13cm,
    height=5cm,
    bar width = 2pt,
    ybar = .05cm,
    ymode=log,
    xmin = 2011,
    xmax = 2023,
    xtick={2012, 2013,..., 2022},
    xticklabels={2012, 2013,..., 2022},
    title = \textbf{(A)},
    ylabel = Number of Publications,
    xlabel = Year,
    enlarge y limits = {abs=0.8},
    legend style={
			at={(0.5,-0.3)},
			anchor=north,
			legend columns=-1},
]

\addplot[red,fill=red] coordinates { (2022, 1379) (2021, 1212) (2020, 1057) (2019, 907) (2018, 808) (2017, 655) (2016, 626) (2015, 421) (2014, 429) (2013, 310) (2012, 300)};

\addplot[red!60,fill=red!60] coordinates { (2022, 188) (2021, 185) (2020, 179) (2019, 161) (2018, 111) (2017, 90) (2016, 59) (2015, 61) (2014, 49) (2013, 53) (2012, 39)};

\addplot[blue,fill=blue] coordinates { (2022, 173) (2021, 121) (2020, 91) (2019, 73) (2018, 40) (2017, 27) (2016, 21) (2015, 10) (2014, 19) (2013, 6) (2012, 13)};

\addplot[blue!60,fill=blue!60] coordinates { (2022, 221) (2021, 122) (2020, 67) (2019, 53) (2018, 40) (2017, 24) (2016, 16) (2015, 25) (2014, 13) (2013, 6) (2012, 7)};

\addplot[blue!30,fill=blue!30] coordinates { (2022, 17) (2021, 23) (2020, 16) (2019, 11) (2018, 9) (2017, 1) (2016, 2) (2015, 3) (2014, 3) (2013, 5) (2012, 1)};


\end{axis}
\end{tikzpicture}
}
\resizebox{0.8\textwidth}{!}{
\begin{tikzpicture}
\begin{axis} [
    width=13cm,
    height=5cm,
    bar width = 3pt,
    ybar = .05cm,
    xmin = 2011,
    xmax = 2023,
    xtick={2012, 2013,..., 2022},
    xticklabels={2012, 2013,..., 2022},
    title = \textbf{(B)},
    ylabel = Publications Percentage,
    xlabel = Year,
    enlarge y limits = {abs=0.8},
    legend style={
			at={(0.5,-0.3)},
			anchor=north,
			legend columns=-1},
]

\addplot[red,fill=red] coordinates { (2022, 69.7) (2021, 72.8) (2020, 74.9) (2019, 75.2) (2018, 80.1) (2017, 82.1) (2016, 86.4) (2015, 80.9) (2014, 83.6) (2013, 81.5) (2012, 83.3)};

\addplot[red!60,fill=red!60] coordinates { (2022, 9.5) (2021, 11.1) (2020, 12.6) (2019, 13.3) (2018, 11.0) (2017, 11.2) (2016, 8.1) (2015, 11.7) (2014, 9.5) (2013, 13.9) (2012, 10.8)};

\addplot[blue,fill=blue] coordinates { (2022, 8.7) (2021, 7.2) (2020, 6.4) (2019, 6.0) (2018, 3.9) (2017, 3.3) (2016, 2.9) (2015, 1.9) (2014, 3.7) (2013, 1.5) (2012, 3.6)};

\addplot[blue!60,fill=blue!60] coordinates { (2022, 11.1) (2021, 7.3) (2020, 4.7) (2019, 4.3) (2018, 3.9) (2017, 3.0) (2016, 2.2) (2015, 4.8) (2014, 2.5) (2013, 1.5) (2012, 1.9)};

\addplot[blue!30,fill=blue!30] coordinates { (2022, 0.8) (2021, 1.3) (2020, 1.1) (2019, 0.9) (2018, 0.8) (2017, 0.1) (2016, 0.2) (2015, 0.5) (2014, 0.5) (2013, 1.3) (2012, 0.2)};


\end{axis}
\end{tikzpicture}
}
\resizebox{0.8\textwidth}{!}{
\begin{tikzpicture}
\begin{axis} [
    width=13cm,
    height=5cm,
    bar width = 3pt,
    ybar = .05cm,
    ymode=log,
    xmin = 2011,
    xmax = 2023,
    xtick={2012, 2013,..., 2022},
    title = \textbf{(C)},
    ylabel = Number of Publications,
    xlabel = Year,
    enlarge y limits = {abs=0.8},
    legend style={
			at={(0.5,-0.3)},
			anchor=north,
			legend columns=-1},
]

\addplot[red,fill=red] coordinates { (2022, 4822) (2021, 4613) (2020, 3983) (2019, 3490) (2018, 3194) (2017, 2875) (2016, 2938) (2015, 2256) (2014, 2022) (2013, 1697) (2012, 1613)};

\addplot[red!60,fill=red!60] coordinates { (2022, 870) (2021, 926) (2020, 926) (2019, 886) (2018, 783) (2017, 732) (2016, 648) (2015, 610) (2014, 543) (2013, 415) (2012, 379)};

\addplot[blue,fill=blue] coordinates { (2022, 758) (2021, 750) (2020, 595) (2019, 540) (2018, 439) (2017, 356) (2016, 329) (2015, 231) (2014, 174) (2013, 168) (2012, 164)};

\addplot[blue!60,fill=blue!60] coordinates { (2022, 517) (2021, 465) (2020, 345) (2019, 239) (2018, 227) (2017, 167) (2016, 137) (2015, 133) (2014, 92) (2013, 86) (2012, 74) };

\addplot[blue!30,fill=blue!30] coordinates { (2022, 80) (2021, 74) (2020, 54) (2019, 32) (2018, 50) (2017, 34) (2016, 32) (2015, 19) (2014, 15) (2013, 17) (2012, 11)};

\legend{Northern Africa, Southern Africa, Western Africa, Eastern Africa, Central Africa}

\end{axis}
\end{tikzpicture}
}
\caption{Scopus-indexed computer vision publications per African region across the time interval 2012-2022. We use the logarithmic scale in (A, C). (A) The number of publications in the refined set. (B) The percentages of publications per region. (C) The number of publications in the \emph{full} set. The two most publishing regions (Northern - Southern Africa) are highlighted in red. It shows consistent growth in Northern and Southern regions’ publications, and a recent increase in Eastern and Western Africa publications in 2016-2022. However, Central Africa is the most in need of improving the computer vision capacity, as it contributes less than 1\% to the total publications.}
\label{fig:geotemporal_refined}
\end{figure}

We start with demonstrating the publishing patterns for the different African regions in the computer vision field over the last ten years (2012-2022). Figure~\ref{fig:geotemporal_refined} (A, B) shows the number of publications and percentages resp., on the logarithmic scale, from the \emph{refined} set with at least one author working in an African institution. Publications with authors affiliated with institutions from two different African countries are accounted for in each country and region, as we are studying the computer vision research capacity. Figure~\ref{fig:geotemporal_refined} (A) shows that institutions in Northern and Southern Africa are the two highest regions publishing in computer vision. They construct around 88.5\% of the total publications in the \emph{refined} set over the last ten years, and exhibit consistent growth over the years. 

Looking upon Eastern and Western Africa we notice growth of the publications over the period 2016-2022. Interestingly, Figure~\ref{fig:geotemporal_refined} (B) shows that the percentage of publications originating from Eastern and Western regions combined has grown from 5.1\% in 2016 to 19.9\% in 2022, closing the gap with Southern Africa. Nonetheless, the gap between sub-Saharan and Northern African remains. More importantly, when looking to Central Africa it is the most in need of building capacity for computer vision research, contributing only 0.9\% of the total publications.

The \emph{refined} set has one fifth of the publications in the \emph{full} set, but has gone through a verification process making it more accurate, as explained in Section~\ref{sec:method}. Nonetheless, we contrast the geo-temporal analysis for both the \emph{refined} and \emph{full} sets, where we show in Figure~\ref{fig:geotemporal_refined} (C) the number of publications in the \emph{full} set for the different African regions over the last ten years. It still shows that the Northern and Southern regions are the highest two regions publishing, with around 85\% from the total of publications stemming from them. It also has Central Africa the least region publishing in computer vision and only contributing around 0.85\%. Thus, it aligns with our previous \emph{refined} set findings.

The previous analysis shows some of the strength points of the computer vision research in Africa such as the development of Western and Eastern African regions to have a better capacity in computer vision research from 2016 till 2022. Moreover, the growth trends in both Northern and Southern regions computer vision publications show promising potential. Yet, our main weakness is the wide gap between the Northern region \textit{vs.} the rest of Africa, especially Central Africa which needs investment in improving their computer vision research. 

\begin{figure}
\begin{minipage}{0.8\linewidth}
    \centering
    \resizebox{\textwidth}{!}{
\begin{tikzpicture}
\begin{axis} [
    width=13cm,
    height=5cm,
    ymode=log,
    xmin = 2011,
    xmax = 2023,
    xtick={2012, 2013,..., 2022},
    title = \textbf{},
    ylabel = Number of Publications,
    xlabel = Year,
    enlarge y limits = {abs=0.8},
    legend style={
			at={(0.5,-0.3)},
			anchor=north,
			legend columns=5},
]

\addplot[line width=1pt, solid, color=red, mark options={scale=0.8, solid}, mark=*] coordinates { (2022, 600) (2021, 530) (2020, 380) (2019, 341) (2018, 224) (2017, 151) (2016, 184) (2015, 126) (2014, 139) (2013, 106) (2012, 120)};

\addplot[line width=1pt, dashed, color=red,mark options={scale=0.8, solid}, mark=*] coordinates { (2022, 261) (2021, 230) (2020, 247) (2019, 167) (2018, 234) (2017, 226) (2016, 176) (2015, 111) (2014, 166) (2013, 86) (2012, 85)};

\addplot[line width=1pt, dotted, color=red,mark options={scale=0.8, solid}, mark=*] coordinates { (2022, 237) (2021, 209) (2020, 179) (2019, 181) (2018, 130) (2017, 99) (2016, 112) (2015, 111) (2014, 78) (2013, 76) (2012, 60)};

\addplot[line width=1pt, solid, color=blue, mark options={scale=0.8, solid}, mark=*] coordinates { (2022, 247) (2021, 212) (2020, 227) (2019, 197) (2018, 190) (2017, 167) (2016, 142) (2015, 65) (2014, 43) (2013, 31) (2012, 33)};

\addplot[line width=1pt, dashed,  color=blue, mark options={scale=0.8, solid}, mark=*] coordinates { (2022, 160) (2021, 158) (2020, 163) (2019, 149) (2018, 100) (2017, 86) (2016, 56) (2015, 57) (2014, 46) (2013, 51) (2012, 38)};

\addplot[line width=1pt, dotted, color=blue, mark options={scale=0.8, solid}, mark=*] coordinates { (2022, 94) (2021, 82) (2020, 50) (2019, 46) (2018, 25) (2017, 18) (2016, 13) (2015, 4) (2014, 12) (2013, 3) (2012, 7)};

\addplot[line width=1pt, solid, color=cyan, mark options={scale=0.8, solid}, mark=*] coordinates { (2022, 148) (2021, 57) (2020, 22) (2019, 20) (2018, 20) (2017, 8) (2016, 4) (2015, 8) (2014, 2) (2013, 0) (2012, 0)};

\addplot[line width=1pt, dashed, color=cyan, mark options={scale=0.8, solid}, mark=*] coordinates { (2022, 53) (2021, 22) (2020, 20) (2019, 15) (2018, 4) (2017, 1) (2016, 2) (2015, 3) (2014, 4) (2013, 1) (2012, 2)};

\addplot[line width=1pt, dotted, color=cyan, mark options={scale=0.8, solid}, mark=*] coordinates { (2022, 13) (2021, 11) (2020, 13) (2019, 16) (2018, 24) (2017, 9) (2016, 7) (2015, 3) (2014, 2) (2013, 8) (2012, 2)};

\addplot[line width=1pt, solid, color=magenta, mark options={scale=0.8, solid}, mark=*] coordinates { (2022, 23) (2021, 16) (2020, 15) (2019, 10) (2018, 6) (2017, 5) (2016, 7) (2015, 6) (2014, 4) (2013, 1) (2012, 1)};

\legend{Egypt, Tunisia, Algeria, Morocco, South Africa, Nigeria, Ethiopia, Ghana, Sudan, Kenya}

\end{axis}
\end{tikzpicture}
}
\end{minipage}%
\begin{minipage}{0.2\linewidth}
\centering
\begin{tabular}{|ll|}
\hline
Egypt  & 3,495 \\
Tunisia &  2,314 \\
Algeria & 1,801 \\
Morocco &  1,706\\
South Africa & 1,396 \\
Nigeria & 379 \\
Ethiopia & 301 \\
Ghana & 131 \\
Sudan & 120 \\
Kenya  & 106 \\
 \hline
\end{tabular}%
\end{minipage}
\caption{Top ten African countries publishing in computer vision from the \textit{refined} set. \textbf{Left:} Number of annual publications in computer vision from the top ten African countries in logarithmic scale. \textbf{Right:} Total number of publications from the top ten African countries.}
\label{fig:top10}
\vspace{-1em}
\end{figure}
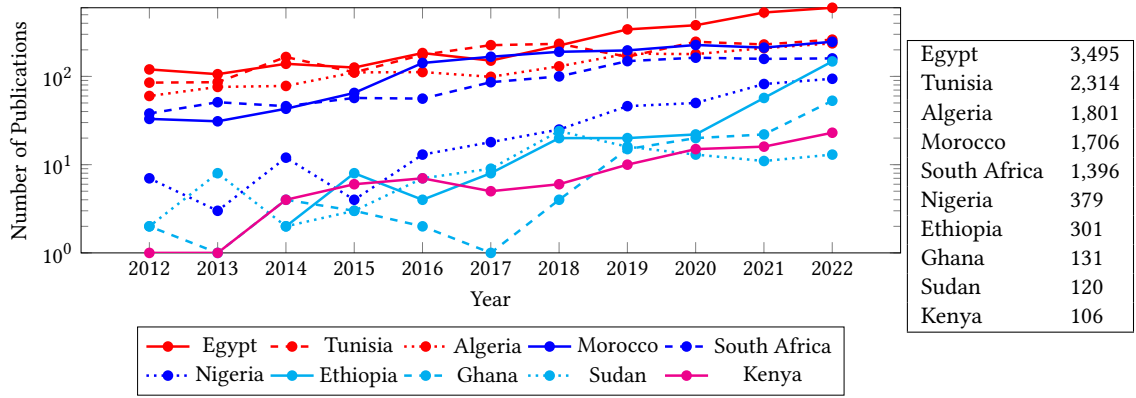

Additionally, we analyze the top ten African countries publishing in computer vision. Figure~\ref{fig:top10} (Left) shows the annual publications for each of the top ten African countries publishing in computer vision, and Figure~\ref{fig:top10} (right) lists these top ten countries with their total publications over the last ten years. They show growth over the years in their number of publications, yet most of them belong to the Northern and Southern regions. Positively, it demonstrates that a country like Ethiopia, outside these two regions, went from zero publications in 2012-2013 to 148 publications in 2022. 

In order to assess the equity in computer vision research and access of African researchers to opportunities and publishing in top-tier venues, we perform a separate geo-temporal analysis on these publications only. We mainly focus on the toughest and highest ranking computer vision conferences (i.e., \emph{CVPR, ICCV and ECCV}), the best machine learning conferences that do accept computer vision publications (i.e., \emph{ICML, NeurIPS and ICLR}), the top medical image processing conference (i.e., \emph{MICCAI}) and the top computer vision journals (i.e., \emph{TPAMI} and \emph{IJCV}). Figure~\ref{fig:toptier} shows the numbers of publications in the aforementioned venues across the different continents over the last ten years. We show the publication-researcher pairs as we want to fairly document the amount of researchers from African institutions that are able to publish in these leading venues. It clearly shows that North America and Asia are the ones mostly publishing in top-tier venues with around three quarters (74\%) of these publications stemming from them. Moreover, they show consistent increase in the publications over the years, especially Asia that surpassed North America in the last two years. However, when looking to Africa it only constitutes 0.06\% of the total publications, and over the years it does not show a consistent increase and growth in the number of publications. Interestingly, we found that Madagascar had two researchers that published in \emph{ICLR} 2022, which shows the promise of how a small under-resourced country was still able to publish in a leading venue as such. Nonetheless, when we compare Africa in general to South America which is another under-resourced continent, we see that South America has more publications where it constitutes around 0.23\% of the total \emph{top-tier} set. This confirms that Africans do not have the same opportunities and access to conduct such research with respect to other continents even when compared to ones that are low resourced. It brings the question of why and what can be done to improve the situation in Africa.

We conduct a similar study but with focus on \emph{Computer Vision and Pattern Recognition (CVPR)} conference only as the well established and frequent venue in computer vision. Additionally, we highlight the African authors affiliated with non African institutions using their affiliation history as a proxy. We identified 49 authors that were affiliated with non African institutions but had an African institution in their affiliation history. Figure~\ref{fig:toptier_cvonly} shows both African authors in African institutions (red) and African authors even if affiliated with non African institutions (blue). It shows worse trends, where Africa constitutes only 0.02\% of the total \emph{CVPR} publications over the last ten years and in certain years no publications (i.e., 2012-2013, 2018-2021). It confirms on the problem as these results are recent, and showing that these issues still persist. It motivates our pursue of first documenting the problem then starting a discussion on why this is the case. However, when we look at the African authors not necessarily affiliated with African institutions, we see a promising potential where their publications constitute 0.2\% of the last ten year \emph{CVPR} publications. This most likely stems from international institutions admitting African graduate students and welcoming them to increase diversity within research and encourage these researchers to pursue their career. Yet, it still begs the question of why the same researchers did not publish in \emph{CVPR} at their African institutions, what can be done to bridge this gap.

\begin{figure}
\begin{minipage}{0.83\linewidth}
    \centering
    \resizebox{\textwidth}{!}{
\begin{tikzpicture}
\begin{axis} [
    width=13cm,
    height=5cm,
    bar width = 2.5pt,
    ybar = .05cm,
    ymode=log,
    xmin = 2011,
    xmax = 2023,
    xtick={2012, 2013,..., 2022},
    title = \textbf{},
    ylabel = Publication-Researcher Pairs,
    xlabel = Year,
    enlarge y limits = {abs=0.8},
    legend style={
			at={(0.5,-0.3)},
			anchor=north,
			legend columns=-1},
]

\addplot[black,fill=black] coordinates { (2022, 7498) (2021, 9926) (2020, 11073) (2019, 9120) (2018, 8620) (2017, 6058) (2016, 4487) (2015, 3813) (2014, 3486) (2013, 3461) (2012, 3030)};

\addplot[black!80,fill=black!80] coordinates { (2022, 14853) (2021, 11948) (2020, 10431) (2019, 7622) (2018, 4925) (2017, 3205) (2016, 2276) (2015, 2183) (2014, 1585) (2013, 1713) (2012, 1402)};

\addplot[black!60,fill=black!60] coordinates { (2022, 4892) (2021, 6060) (2020, 5962) (2019, 4860) (2018, 4318) (2017, 3319) (2016, 2432) (2015, 2358) (2014, 2006) (2013, 2297) (2012, 2095)};

\addplot[black!40,fill=black!40] coordinates { (2022, 1024) (2021, 803) (2020, 796) (2019, 611) (2018, 491) (2017, 378) (2016, 299) (2015, 284) (2014, 260) (2013, 267) (2012, 164)};

\addplot[black!20,fill=black!20] coordinates { (2022, 40) (2021, 83) (2020, 69) (2019, 32) (2018, 48) (2017, 21) (2016, 21) (2015, 39) (2014, 21) (2013, 27) (2012, 13)};

\addplot[red,fill=red] coordinates { (2022, 21) (2021, 9) (2020, 38) (2019, 8) (2018, 11) (2017, 3) (2016, 5) (2015, 2) (2014, 9) (2013, 2) (2012, 7)};

\legend{North America, Asia, Europe, Oceania, South America, Africa}

\end{axis}
\end{tikzpicture}
}
\end{minipage}%
\begin{minipage}{0.17\linewidth}
\centering
\begin{tabular}{|ll|}
\hline
SA & 48 \\
Egypt &  20\\
Algeria & 11 \\
Morocco & 8 \\
Rwanda & 8 \\
Tunisia & 8 \\
Kenya & 4\\
Ghana & 2\\
Madagascar & 2\\
Nigeria & 2\\
Uganda & 2\\
 \hline
\end{tabular}%
\end{minipage}
\caption{Publications in top-tier venues (\emph{CVPR, ICCV, ECCV, ICML, NeurIPS, ICLR, MICCAI, TPAMI, IJCV}) across all continents. \textbf{Right:} Number of researcher-publication pairs in top-tier venues per continent over the last ten years, with Africa highlighted in red. \textbf{Left:} The top African countries publishing in top-tier venues. SA: South Africa.}
\label{fig:toptier}
\vspace{-2em}
\end{figure}
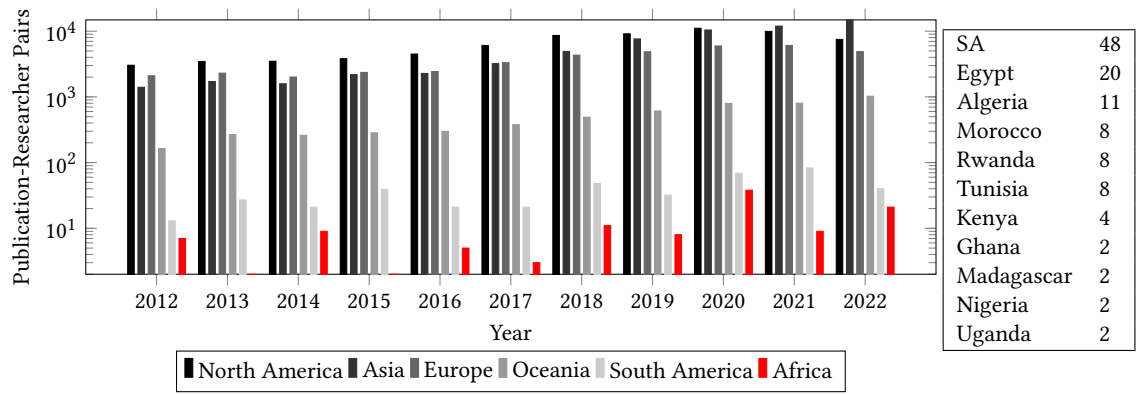
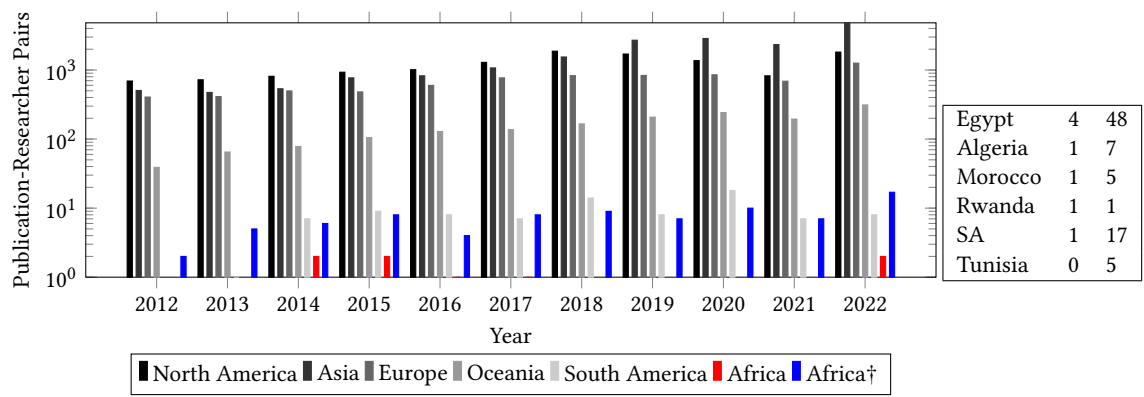
\begin{figure}
\begin{minipage}{0.83\linewidth}
    \centering
    \resizebox{\textwidth}{!}{
\begin{tikzpicture}
\begin{axis} [
    width=13cm,
    height=5cm,
    bar width = 2pt,
    ybar = .05cm,
    ymode=log,
    xmin = 2011,
    xmax = 2023,
    xtick={2012, 2013,..., 2022},
    title = \textbf{},
    ylabel = Publication-Researcher Pairs,
    xlabel = Year,
    enlarge y limits = {abs=0.8},
    legend style={
			at={(0.5,-0.3)},
			anchor=north,
			legend columns=-1},
]

\addplot[black,fill=black] coordinates { (2022, 1819) (2021, 824) (2020, 1368) (2019, 1705) (2018, 1875) (2017, 1292) (2016, 1010) (2015, 930) (2014, 812) (2013, 722) (2012, 693)};

\addplot[black!80,fill=black!80] coordinates { (2022, 4820) (2021, 2334) (2020, 2862) (2019, 2704) (2018, 1544) (2017, 1073) (2016, 824) (2015, 770) (2014, 535) (2013, 474) (2012, 506)};

\addplot[black!60,fill=black!60] coordinates { (2022, 1259) (2021, 689) (2020, 852) (2019, 835) (2018, 828) (2017, 770) (2016, 599) (2015, 482) (2014, 498) (2013, 412) (2012, 406)};

\addplot[black!40,fill=black!40] coordinates { (2022, 312) (2021, 195) (2020, 242) (2019, 208) (2018, 166) (2017, 137) (2016, 129) (2015, 105) (2014, 78) (2013, 65) (2012, 39)};

\addplot[black!20,fill=black!20] coordinates { (2022, 8) (2021, 7) (2020, 18) (2019, 8) (2018, 14) (2017, 7) (2016, 8) (2015, 9) (2014, 7) (2013, 1) (2012, 0)};

\addplot[red,fill=red] coordinates { (2022, 2) (2021, 0) (2020, 0) (2019, 0) (2018, 0) (2017, 1) (2016, 1) (2015, 2) (2014, 2) (2013, 0) (2012, 0)};

\addplot[blue,fill=blue] coordinates { (2022, 17) (2021, 7) (2020, 10) (2019, 7) (2018, 9) (2017, 8) (2016, 4) (2015, 8) (2014, 6) (2013, 5) (2012, 2)};
\legend{North America, Asia, Europe, Oceania, South America, Africa, Africa$\dagger$}

\end{axis}
\end{tikzpicture}
}
\end{minipage}%
\begin{minipage}{0.17\linewidth}
\centering
\begin{tabular}{|lll|}
 \hline
Egypt &  4 & 48\\
Algeria & 1 & 7\\
Morocco & 1 & 5\\
Rwanda & 1 & 1\\
SA & 1 & 17\\
Tunisia & 0 & 5 \\
 \hline
\end{tabular}%
\end{minipage}
\caption{Publications in the best computer vision venue (\emph{CVPR}). \textbf{Right:} Number of researcher-publication pairs in top-tier venues per continent over the last ten years, Africa highlighted in red.  Africa$\dagger$: indicates African authors not necessarily affiliated with African institutions and highlighted in blue. \textbf{Left:} The top African countries publishing in \emph{CVPR}. SA: South Africa. Table Column I: The number of pairs having authors affiliated with African institutions, Table Column II: The number of pairs having African Authors not necessarily affiliated with African institutions.}
\label{fig:toptier_cvonly}
\end{figure}

\subsection{Collaboration patterns analysis}
\begin{figure}
\centering
\resizebox{0.8\textwidth}{!}{
\begin{tikzpicture}
\begin{axis} [
    width=13cm,
    height=5cm,
    bar width = 4pt,
    ybar = .05cm,
    ymode=log,
    xmin = 2010,
    xmax = 2023,
    xtick={2011, 2012, 2013,..., 2022},
    xticklabels={pre2012, 2012, 2013,..., 2022},
    title = \textbf{(A)},
    ylabel = Number of Publications,
    xlabel = Year,
    enlarge y limits = {abs=0.8},
    legend style={
			at={(0.5,-0.3)},
			anchor=north,
			legend columns=-1},
]

\addplot[black!50,fill=black!50] coordinates { (2022, 1809) (2021, 1128) (2020, 891) (2019,656) (2018, 519) (2017, 332) (2016, 329) (2015, 256) (2014, 267) (2013, 164) (2012, 181) (2011, 851)};

\addplot[red,fill=red] coordinates { (2022, 88) (2021, 57) (2020, 45) (2019, 24) (2018, 25) (2017, 9) (2016, 8) (2015, 9) (2014, 7) (2013, 2) (2012, 5) (2011, 21)};

\legend{International, African}

\end{axis}
\end{tikzpicture}
}
\resizebox{0.8\textwidth}{!}{
\begin{tikzpicture}
\begin{axis} [
    width=13cm,
    height=5cm,
    bar width = 4pt,
    ybar = .05cm,
    ymode=log,
    xmin = 2010,
    xmax = 2023,
    xtick={2011, 2012, 2013,..., 2022},
    xticklabels={pre2012, 2012, 2013,..., 2022},
    title = \textbf{(B)},
    ylabel = Number of Publications,
    xlabel = Year,
    enlarge y limits = {abs=0.8},
    legend style={
			at={(0.5,-0.3)},
			anchor=north,
			legend columns=-1},
]
											
\addplot[red,fill=red] coordinates { (2022, 1347) (2021, 1304) (2020, 1162) (2019, 970) (2018, 817) (2017, 675) (2016, 607) (2015, 435) (2014, 411) (2013, 319) (2012, 291) (2011, 1259)};

\addplot[red!50,fill=red!50] coordinates { (2022, 1335) (2021, 1138) (2020, 989) (2019, 809) (2018, 712) (2017, 555) (2016, 499) (2015, 340) (2014, 344) (2013, 259) (2012, 217) (2011, 1019)};
											
\addplot[black!40,fill=black!40] coordinates { (2022, 1383) (2021, 1161) (2020, 1032) (2019, 840) (2018, 734) (2017, 588) (2016, 522) (2015, 355) (2014, 364) (2013, 263) (2012, 236) (2011, 1111)};
											
\legend{First, Last, Others}

\end{axis}
\end{tikzpicture}
}
\caption{Collaboration patterns analysis. (A) The distribution of African versus international collaborations over the last ten years. (B) The distribution of African authors as a key contributor to the work (i.e., first or last authors) \textit{vs.} other roles.}
\label{fig:collabs_africa_vs_international}
\vspace{-2em}
\end{figure}
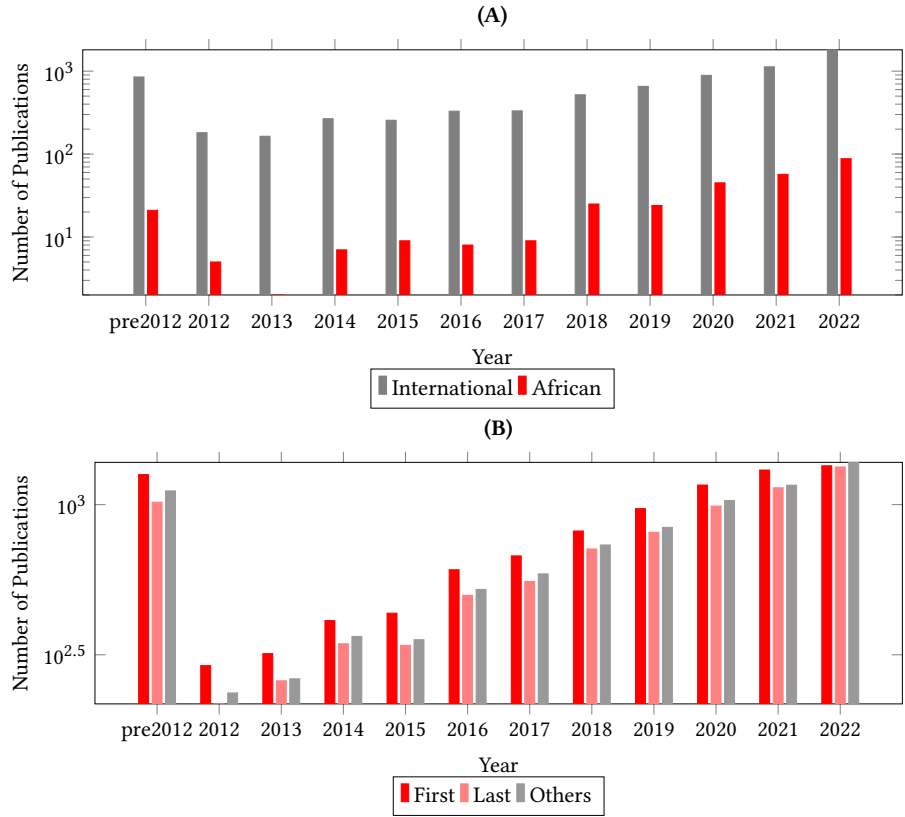

In order to understand the collaboration patterns in African computer vision publications, we conduct two types of analysis on the \emph{refined} set. First we analyze the African \textit{vs.} international collaborations in terms of the number of publications over the last ten years. Figure~\ref{fig:collabs_africa_vs_international} (A) demonstrates that most of the publications are dominated by international collaborations with very minimal amount of African collaborations~\cite{pouris2014research,turki2023machine}, forming only 3.9\% of the total publications. Since African countries share some of the problems and bottlenecks they are facing, pan-African collaborations can strengthen the continent's research eco-system. Moreover, it can result in a better focus on our own problems, and will make the reciprocal engagements between the affected communities and researchers easier as the researchers belong to these affected communities. Nonetheless, we think that international collaborations does improve African research in other aspects, because of the fact that the continent is limited in resources (e.g. compute). These collaborations can help overcome these barriers. Hence, why we believe that a balance between pan-African and international collaborations is necessary towards an independent African computer vision research eco-system. We also list here the top ten countries outside Africa that act as international collaborators: France (28\%), Saudi Arabia (16.6\%), United States (14.3\%), India (8.8\%), China (8.2\%), United Kingdom (7.9\%), Canada (5.3\%), Germany (4.7\%), Spain (3.8\%), and United Arab Emirates(2.4\%). We list the countries with the percentages of publications for each. 

Second, we analyze the role of African authors in such publications, whether they are a key-contributor (i.e., first or last author) or rather a co-author that is not a key contributor. In this analysis publications that have same African country with differing roles (e.g., first and last) are counted twice in each role. Figure~\ref{fig:collabs_africa_vs_international} (B) shows the amount of publications per year with first, last African authors or others. It clearly shows a good strength point in our community, where most of the publications have African authors as key contributors in the work either as first or last authors.

\subsection{Keywords analysis}
In our final analysis, we study the research topics and recurring keywords in the computer vision field in Africa on the \emph{full} set of publications. In this analysis we choose to do it using the \emph{full} instead of the \emph{refined} set, since the \emph{refined} set depends on the top 50 keywords from global computer vision in their query generation. Thus, it can skew the results. These keywords are retrieved using an off-the-shelf tool as described in Section~\ref{sec:method_details}. Out of 187,812 keywords we only identify the top-30 recurring ones. In Figure~\ref{fig:plot_highly_cited} (A-E), we show these top-30 keywords per African region, to identify the topics distribution per region. We also show in Figure~\ref{fig:plot_highly_cited} (F) the co-occurrence matrix between pairs of these keywords, which can be used to understand certain keywords that non experts in the topic might not be aware of. For example, \emph{Hyperchaotic Systems} is highly co-occurring with \emph{Image Encryption}, where we see that it does relate to techniques in cryptography. 
Surprisingly, we found keywords such as \emph{Galaxies} and \emph{Crystalline Texture}, to verify their relevance we inspect five random publications for each. We found that \emph{Crystalline Texture} is used in publications related to texture classification which is relevant to computer vision. However, the keyword \emph{Galaxies} is used in publications that are relevant~\cite{fielding2022classification}, but others are not~\cite{shirley2021help}.

Inspecting the top 30 keywords distribution per keyword and African region, Figure~\ref{fig:plot_highly_cited} (A) shows that for \emph{Image Segmentation} Northern Africa contributes higher than other regions with around 90\%. Interestingly, Figure~\ref{fig:plot_highly_cited} (B) shows that \emph{Galaxies} is mostly researched in Southern Africa, which is hard to discern automatically whether it is relevant to computer vision or not as detailed earlier. Yet we identify some publications that are categorized under Computer Vision and it might be related to the Square Kilometre Array project that is hosted in Southern Africa. 
Figure~\ref{fig:plot_highly_cited} (C, D) show both Eastern and Western Africa with keywords \emph{Landsat} and \emph{Land Cover} showing as around the second or third regions researching that topic, where Northern and Southern regions are mostly dominating these. 
The question of whether certain regions are working on their most urgent needs or not remains unanswered but is beyond the scope of this study. Nonetheless, this distribution of topics per region is an enabler for researchers and policy makers to make informed decisions on this previous question.

\subsection{Summary and Discussion}
In this section, we summarize our results and list open questions. Our analysis has shown that African researchers are key contributors to African research, yet there exists multiple barriers facing Africans across all regions to publish in top-tier venues. We showed African contributions constituting only 0.06\% over the last ten years. It brings the question of, \textbf{what could be these potential barriers and how to improve our research eco-system to overcome them?} It also showed the current trend of topics published in Africa, bringing the question whether our research is addressing our most urgent needs. \textbf{Is \emph{Image Encryption} the most needed topic in Northern or Central Africa?} Moreover, it showed the current gap in Central Africa region with respect to other regions, and the dominance of Northern region in computer vision publications. It still showed a recent increase of research output from Eastern and Western Africa. Nonetheless, it brings the question of \textbf{how to increase the computer vision research capacity in Africa and especially Central Africa?} 

In order to discuss the first open question within a participatory framework, we ask 14 community members from Egypt, Nigeria, Cameroon and Benin about the barriers facing African computer vision research, suggestions to improve it and its strengths. We ensure to take their informed consent. Note, that this is a discussion to incorporate our own stories in the work to lead the way to a large-scale quantitative study for our future work. The responses from the community members showed a wide agreement that some of the main barriers in Africa~\cite{de2018machine} include, lack of funding, limited availability of data, poor economic systems and low access to resources. Another barrier is the existence of peer-reviewing bias in venues as previously documented~\cite{tomkins2017reviewer}. Additionally, the responses showed that there can be a disconnect between academic research and industry, which might be attributed to lack of well established datasets collected from the continent, and lack of incentives for companies and startups to conduct research. Another barrier is the unreasonable teaching loads on the graduate students and low compensation for academics and researchers. Low incentives from institutions to conduct high quality research. There is also lack of in-depth technical knowledge in computer vision in Africa when compared to general machine learning. Moreover, the low quality of education in some of the African institutions makes the students struggle. Finally, the low presence of senior researchers or academics who can provide mentorship and supervision is also a contributing barrier, since African researchers tend to travel outside Africa. Some of the strengths in the African countries, as a general consensus from the community, is the presence of many computer vision problems (e.g. healthcare, education, etc.) that are specifically relevant to Africa. These diverse problems open the door for significant contributions from African researchers. Moreover, the abundance of passionate African researchers that want to help their communities. Accordingly, there is a need for funding, collaborations between academia and industry, access to training and resources to engage these passionate researchers in their communities' needs. 

\input{graphs/keywork_regional}

\section{Conclusion}
We presented a bibliographic analysis of the computer vision publications in Africa to document inequity in computer vision research among the different regions and globally. For our future work, we aim to start three initiatives: (i) A quantitative survey on the structural barriers facing African computer vision researchers. (ii) The creation of an academic committee to discuss computer vision syllabus. (iii) Study of the biases in top-tier venues reviewing process.

\begin{acks}
We thank \emph{SisonkeBiotik} and \emph{Masakhane} grassroots for their support in the development of this research paper. 
\end{acks}

\bibliographystyle{ACM-Reference-Format}
\bibliography{references}

\end{document}